\documentclass[a4paper]{article}

\usepackage{INTERSPEECH2019}
\usepackage{amsmath,amssymb,amsfonts}
\usepackage{graphicx}
\usepackage{tikz}
\usetikzlibrary{shapes, calc, shapes, arrows,fit, decorations.pathreplacing}
\usepackage{textcomp}
\usepackage{xcolor}
\newcommand{\symbolvec}[1]{\vec{\mathbf{#1}}}
\newcommand{\symbolmatrix}[1]{\mathbf{#1}}

\title{Practical applicability of deep neural networks for overlapping speaker separation}
\name{Pieter Appeltans$^1$, Jeroen Zegers$^2$, Hugo Van hamme$^2$}
\address{
  $^1$Department of Computer Science - KU Leuven - Belgium\\
  $^2$Processing Speech and Images - ESAT - KU Leuven - Belgium}
\email{\{pieter,jeroen,hugo\}.\{appeltans,zegers,vanhamme\}@kuleuven.be}

\begin{document}
\maketitle
\begin{abstract}
This paper examines the applicability in realistic scenarios of two deep learning based solutions to the overlapping speaker separation problem. Firstly, we present experiments that show that these methods are applicable for a broad range of languages. Further experimentation indicates limited performance loss for untrained languages, when these have common features with the trained language(s). Secondly, it investigates how the methods deal with realistic background noise and proposes some modifications to better cope with these disturbances. The deep learning methods that will be examined are deep clustering and deep attractor networks.
\end{abstract}
\noindent\textbf{Index Terms}: Source Separation, Recurrent neural networks, Artificial neural networks  

\section{Introduction}
The overlapping speaker separation problem consists of separating the utterances of multiple speakers from a mixture. Many cues, such as the identity, position, and lip movement of the speakers could be used to tackle this problem. This paper however will focus on methods that only use a mono recording of the mixture.  

The current state-of-the-art methods to address the speaker separation problem are based on deep neural networks. These methods are able to obtain good text-independent separations with no or limited prior information \cite{deep_clustering}. This is a big improvement compared to previous methods like hidden Markov models \cite{HERSHEY201045,source_adaption,Non_negative_hiden_markov}, independent component analysis \cite{ica_speech}, computational auditory scene analysis \cite{hu2013unsupervised,wang2006computational} and non-negative matrix factorisation \cite{nnmf_half_baken_well_done,schmidt_speech_nnmf}, which have limited separation performance or impose restrictions on the speakers and vocabulary. This improved performance comes at the cost of needing a lot of labelled training data (mixtures for which the desired separation is known) and demanding computations. The former can be tackled by artificially generating mixtures from two separate sources. The latter becomes feasible due to the increasing parallel computation power of graphical processing units.

This paper will focus on two such methods, namely deep clustering (DC) \cite{deep_clustering} and deep attractor networks (DAN) \cite{deepattractornet}. Both use (bidirectional)  recurrent neural networks with long-short term memory (LSTM) cells to map each bin in the log magnitude spectrogram of the mixture to an embedding vector. This mapping is learned from training data and is such that embedding vectors associated with bins dominated by the same speaker are close. These vectors are then used to generate masks to filter out the individual speakers from the mixture. By using these intermediate embedding vectors instead of directly outputting the masks, the so called permutation problem \cite{deepattractornet} is avoided. 

This paper is organised as follows. In the remainder of this introduction the specific details of the two examined method are further discussed. Section \ref{sec:different_languages} presents experiments to asses their performance for six different languages. Subsequently, Section \ref{sec:unseen_languages} will examine how well a model trained for one language generalises to another language and how this generalisation changes when multiple languages are used for training. Section \ref{sec:noise} discusses the applicability of these methods in the presence of background noise and proposes some modifications to improve their performance. Finally, Section~\ref{sec:conclusion} gives some overall conclusions.
\subsection{Deep clustering \cite{deep_clustering}}
\label{sec:deep_clustering}
In DC,  the network is trained by minimising the following loss function: 
\begin{equation}
\label{eq:deep_clustering_kost}
\sum_{n=1}^{N} \frac{1}{K_n^2} ||\symbolmatrix{V}_n\symbolmatrix{V}_n^T-\symbolmatrix{Y}_n\symbolmatrix{Y}_n^T||_F^2
\end{equation}
with $N$ the number of mixtures in the training set, $K_n$ the number of time-frequency bins in the spectrogram of mixture $n$, $\symbolmatrix{V}_n$ a $K_n \times D$ (with $D$ the size of the embedding vectors) dimensional matrix with the embedding vectors, outputted by the network, each normalised to (euclidean) norm 1,  $\symbolmatrix{Y}_n$ a $K_n\times C_n$ (with $C_n$ the number of speaker in mixture $n$) dimensional matrix with $y_{(t,f),c} = 1$ if speaker $c$ dominates the time-frequency bin and $y_{(t,f),c} = 0$ else. This cost function can be understood as follows.

$\symbolmatrix{Y}_n\symbolmatrix{Y}_n^T$ and $\symbolmatrix{V}_n\symbolmatrix{V}_n^T$ are $K_n \times K_n$ dimensional matrices. $\{\symbolmatrix{Y}_n\symbolmatrix{Y}_n^T\}[(t,f),(t',f')]$ is equal to one when time-frequency bins $(t,f)$ and $(t',f')$ are dominated by the same speaker and zero in the other case and the $((t,f),(t',f'))$\textsuperscript{th} element of $\symbolmatrix{V}_n\symbolmatrix{V}_n^T$ is the euclidean inner product of $\symbolvec{v}^{(t,f)}$ and $\symbolvec{v}^{(t',f')}$. Minimising the cost function will thus tend to map embedding vectors associated with time-frequency bins dominated by the same speaker near each other (${\symbolvec{v}^{(t,f)}}^{T}\symbolvec{v}^{(t',f')} \approx 1$) and vectors associated with different speakers will tend to be orthogonal (${\symbolvec{v}^{(t,f)}}^{T}\symbolvec{v}^{(t',f')} \approx 0$). 

After training, the network is used to separate new unseen mixtures. This is done by applying its log magnitude spectrogram to the network and clustering the resulting embedding vectors with K-means. Each cluster represents one speaker and is used to create a binary mask to reconstruct the original utterance of the speaker.
\subsection{Deep attractor networks \cite{deepattractornet}}
\label{sec:deep_attractor}
In DANs the network is trained by minimising: 
\begin{equation}
\label{eq:deep_attractor_networks_loss}
\sum_{n=1}^{N} \frac{1}{K_n*C_n}\sum_{c=1}^{C_n}||\symbolmatrix{S}^{mag}_{n,c} - \symbolmatrix{X}^{mag}_n \odot \symbolmatrix{M}_{n,c}||_F^2
\end{equation}
with $\symbolmatrix{X}^{mag}_n$ the magnitude spectrogram of mixture $n$, $\symbolmatrix{S}^{mag}_{n,c}$ the original magnitude spectrogram of speaker c in mixture $n$, $\odot$ the element wise product, and $\symbolmatrix{M}_{n,c}$ the estimated mask for speaker $c$ that is obtained as follows from the output of the network:
\begin{equation}
\label{eq:dan_mask_sigmoid}
M_{n,c}[t,f] = \frac{1}{1+\exp(- \symbolvec{a}_{c} \cdot \symbolvec{v}(t,f))} 
\end{equation}
with $\symbolvec{a}_{c}$ the attraction point of speaker $c$, which is calculated as the mean of the embedding vectors associated with the speaker:
\begin{equation}
\label{eq:dan_attractor_point}
\symbolvec{a}_{c} = \frac{\sum_{(t,f)}\symbolvec{v}(t,f) y_{(t,f),c}}{\sum_{(t,f)}y_{(t,f),c}} 
\end{equation}
By minimising the above mentioned loss function, the network learns to form an attraction point in embedding space for each speaker, that attracts embedding vectors associated with time-frequency bins of this source. 

To separate an unseen mixture, its log magnitude spectrogram is fed to the network and the obtained embedding vectors are used to create a ratio mask for each speaker using Eq.~(\ref{eq:dan_mask_sigmoid}). Because the partitioning of the bins ($y_{(t,f),c}$) is not known (this is exactly what we are looking for), Eq.~(\ref{eq:dan_attractor_point}) cannot be used to calculate the attraction points. These are therefore approximated by the cluster centres found by K-means clustering of the embedding vectors.
\section{Different languages}

\label{sec:different_languages}
In \cite{deep_clustering} and \cite{deepattractornet} the separation performance of the above mentioned methods is only examined for mixtures of English speakers. This section presents experiments with six other languages, including a tonal language. It is structured as follows: first the experiment design is explained; subsequently the separation scores are presented and discussed.
\subsection{Experiment set-up}
The mixtures are generated using the global phone corpus \cite{schultz2002globalphone} by overlaying utterances of two different speakers. To compare with the results in \cite{deep_clustering} and \cite{deepattractornet}, we used a similar set-up: the signals were subsambled to 8kHz (to limit memory requirements and computation time); we calculate the (log magnitude) spectrogram using the short time Fourier transformation with a cosine window of 32 milliseconds and an overlap of 8 milliseconds; the neural network consisted of two layers of 600 bidirectional LSTM cells, followed by a fully connected layer of neurons with linear activation function; a 20 dimensional embedding space was used. For each language the training set consisted of 20~000 training mixtures, which each contained 2 speakers randomly sampled from a pool of 70 speakers, the development set 3~000 mixtures sampled from 10 speakers and the test set 3~000 mixtures sampled from 20 speakers. The speakers in the different data sets are non-overlapping and in each set there were as many male as female speakers.

The quality of the separations is quantified by the signal to distortion ratio (SDR) which measures the retrieved source energy relative to the energy of interfering sources and artifacts.
\subsection{Results}
Table \ref{tab:dc_dan_verschillende_talen} gives the average SDR for DC and DAN for mixtures of two speakers in respectively Arabic, French, Mandarin, Portuguese, Spanish, and Swedish. These scores are in line to with the results in \cite{deep_clustering} and \cite{deepattractornet} for English. In our experiments deep attractor networks outperform deep clustering for every language and therefore seems the better choice. Both methods obtain their best score for Mandarin, which is the only tonal language in our test set. This might indicate that tonality is a useful feature for speaker separation but more research with other tonal languages is needed to support this thesis. 
\begin{table}[th]
	\centering
	\renewcommand*{\arraystretch}{1.05}
	\caption{The average SDR (in dB) when trained and tested on the same language.}
	\label{tab:dc_dan_verschillende_talen}
	\begin{tabular}{l cc}
		\toprule
		\textbf{language} & \textbf{deep clustering} & \textbf{deep attractor networks}  \\ \midrule
		Arabic & 7.50 & 7.97 \\
		French & 7.46 & 8.20 \\
		Mandarin & 8.54 & 8.86 \\[0.5ex]
		Portuguese & 7.24 & 8.27\\
		Spanish & 6.72 & 7.76 \\
		Swedish & 6.93 & 7.83 \\
		\bottomrule
	\end{tabular}
\end{table}
\section{Generalisation to an unseen language}
\label{sec:unseen_languages}
This section will examine how well a network can separate mixtures of an untrained language. The reasons for these experiments are threefold.
Firstly, it may not be reasonable to assume the speaker's language is know, e.g. when deploying a conferencing service over the internet or when built into a mobile phone.
Secondly, these results give an indication of the robustness against different accents and dialects of a language. Lastly, they might give some information on what cues, such as phonetic, phonotactic, lexical or grammatical, the methods exploit to separate speakers. In Section~\ref{subsec:design_experiment} the set-up of the experiments is described. Next, experiments with networks trained with one language are presented in Section~\ref{sec:network_one_language}. Section~\ref{sec:multiple_languages} examines whether the performance for trained and untrained languages improves when more than one training language is used. 
\subsection{Experimental set-up}
\label{subsec:design_experiment}
In Section~\ref{sec:network_one_language} we reuse the networks from Section \ref{sec:different_languages} trained with respectively French and Swedish speakers. The French network is tested for mixtures with respectively Portuguese and Mandarin speakers. The Swedish network is tested with mixture of respectively Arabic and Spanish speakers. In Section~\ref{sec:multiple_languages} new networks are trained with \{French, Turkish\}, \{French, Turkish, Japanese\}, \{Swedish, Turkish\} and \{Swedish,  Turkish, Japanese\} datasets. For each network the training set consisted of 20 000 two-speaker mixtures sampled from a pool of 70 speakers, equally balanced between languages and genders. The development set consisted of 3 000 mixtures sampled from 10 speakers. Mixtures consisted only of speakers of the same language. The test sets are the same as in Section \ref{sec:different_languages}.
\subsection{Network trained with one language}
\label{sec:network_one_language}
Table \ref{tab:dan_generalisatie_taal} gives the average separation performance of the methods for untrained languages. Also the difference with the score of the network trained with the considered language (Table \ref{tab:dc_dan_verschillende_talen}) is given. 

We noticed that for all languages there is a significant decrease in separation quality compared to the network trained with the test language itself. Portuguese, Arabic and Spanish have a decrease of about 1dB, and the obtained separation are still of good quality. For Mandarin on the other hand the decrease is more significant, around 4dB. This seems to indicate that the performance for untrained languages depends on the relation of the training and test language (closer related is better). 

The fact that the methods do not break completely implies that they do not create grammatical or lexical models, but at most phonotactic or phonetic models. They do seem to do more than tracking formants or pitch, which would make them almost language independent.
\begin{table}[th]
	\centering
	\renewcommand*{\arraystretch}{1.05}
	\caption{The average SDR in dB for DC and DAN for an unseen test language and the difference with the SDR for matched language training.}
	\label{tab:dan_generalisatie_taal}
	\begin{tabular}{l cc}
		\toprule
		\textbf{language} & \textbf{deep clustering} &\textbf{deep attractor networks} \\ \midrule
		& \multicolumn{2}{c}{French network} \\ 
		Mandarin & 4.59 ~ (-3.95)&  4.86 ~ (-4.00) \\
		Portuguese & 6.33 ~ (-1.13) &  7.22 ~ (-0.98)\\[0.5ex]
		& \multicolumn{2}{c}{Swedish network} \\ 
		Arabic & 5.98 ~ (-1.52) &  7.01 ~ (-0.96)\\
		Spanish & 6.01 ~ (-0.71)& 7.20 ~ (-0.55) \\ \bottomrule
	\end{tabular}
	
\end{table}
\subsection{Network trained with multiple languages}
\label{sec:multiple_languages}
Table~\ref{tab:multi_generalisatie_taal} gives the separation quality for the networks trained with multiple languages. The average SDR is reported for both trained (t) languages and untrained (u) languages and the difference with scores of the networks trained with the language itself (Table~\ref{tab:dc_dan_verschillende_talen}). From the results we observe that for trained languages it is in most cases disadvantageous to replace a part of the training data with mixtures in other languages. For untrained languages on the other hand, it is in some cases advantageous to include multiple training languages instead of one. Only for Portuguese there is a consistent decrease in performance compared to the results of the previous subsection.

\begin{table}[tbh]
	\centering
	\renewcommand*{\arraystretch}{1.05}
	\caption{The average SDR in dB for DC and DAN trained with multiple languages for trained and untrained languages and the difference with the SDR for matched language training.}
	\label{tab:multi_generalisatie_taal}
	\begin{tabular}{l cc}
		\toprule
		\textbf{language} & \textbf{deep clustering} & \textbf{deep attractor networks}\\ \midrule
		& \multicolumn{2}{c}{ \small \{French, Turkish\} network} \\ 
		French (t) & 6.92 ~ (-0.55)  & 7.75 ~ (-0.45) \\
		Mandarin (u) & 4.89 ~ (-3.65) &  5.32 ~ (-3.54)  \\ 
		Portuguese (u) & 6.31 ~ (-1.15) & 7.24 ~ (-0.96) \\ [0.5ex]
		& \multicolumn{2}{c}{\small \{French, Turkish, Japanese\} network} \\ 
		French (t) & 6.35 ~ (-1.11) & 7.27 ~ (-0.93) \\
		Mandarin (u) & 4.57 ~ (-3.97)& 5.34 ~ (-3.52) \\ 
		Portuguese (u) & 5.94 ~ (-1.53) & 6.77 ~ (-1.44)\\ [0.5ex]
		& \multicolumn{2}{c}{\small \{Swedish, Turkish\} network} \\ 
		Swedish (t)& 7.03 ~ (0.10) & 6.97 ~ (-0.87)\\
		Arabic (u) & 6.45 ~ (-1.02)& 6.71 ~ (-1.26)\\
		Spanish (u) & 6.39 ~ (-0.33) & 6.99 ~ (-0.77) \\ [1ex]
		& \multicolumn{2}{c}{\small \{Swedish, Turkish, Japanese\} network} \\ 
		Swedish (t) & 6.75 ~ (-0.18) &  7.58 ~ (-0.25)\\
		Arabic (u) & 6.49 ~ (-1.02)& 7.31 ~ (-0.66)\\
		Spanish (u) & 6.16 ~ (-0.56)  & 7.34 ~ (-0.42)  \\ \bottomrule
	\end{tabular}
\end{table}

\section{Coping with background noise}
\label{sec:noise}
In this section we examine the usability of deep clustering and deep attractor networks in the presence of realistic background noise and propose some modifications. This section is organized as follows. First, the modifications to the original methods are presented. Subsequently, the set-up of the experiments is discussed. To conclude, the performance of the original and modified methods are compared. 

\subsection{Proposed modifications}
\label{sec:deep_embedding_noise}
\subsubsection{Modified network architecture}
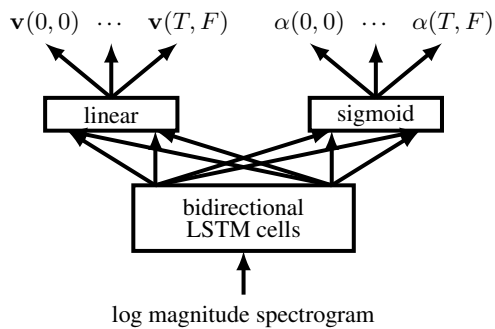
\begin{figure}[th]
	\centering
	\begingroup
	\tikzset{every picture/.style={scale=0.55}}%
\ifx\du\undefined
  \newlength{\du}
\fi
\setlength{\du}{15\unitlength}
\begin{tikzpicture}
\pgftransformxscale{1.000000}
\pgftransformyscale{-1.000000}
\definecolor{dialinecolor}{rgb}{0.000000, 0.000000, 0.000000}
\pgfsetstrokecolor{dialinecolor}
\definecolor{diafillcolor}{rgb}{0.000000, 0.000000, 0.000000}
\pgfsetfillcolor{diafillcolor}
\pgfsetlinewidth{0.100000\du}
\pgfsetdash{}{0pt}
\pgfsetdash{}{0pt}

\pgfsetmiterjoin
\draw (4.0\du,19.0\du)--(4.0\du,22.0\du)--(14.0\du,22.0\du)--(14.0\du,19.0\du)--cycle;

\node at (9.0\du,20.0\du){bidirectional};
\node at (9.0\du,21.0\du){LSTM cells};

\pgfsetbuttcap
{
\pgfsetarrowsend{latex}
\draw (9.0\du,24.0\du)--(9.0\du,22.0\du);
}
\node at (9.000000\du,25.000000\du){log magnitude spectrogram};

\pgfsetbuttcap
{
\pgfsetarrowsend{latex}
\draw (3.0\du,15.0\du)--(0.0\du,12.50000\du);
}

\pgfsetbuttcap
{
\pgfsetarrowsend{latex}
\draw (3.0\du,15.0\du)--(3.0\du,12.50000\du);
}

\pgfsetbuttcap
{
\pgfsetarrowsend{latex}
\draw (3.0\du,15.0\du)--(6.0\du,12.50000\du);
}

\node at (0.0\du,11.6\du){$\symbolvec{v}(0,0)$};
\node at (3.0\du,11.6\du){\dots};
\node at (6.500000\du,11.6\du){$\symbolvec{v}(T,F)$};

\pgfsetmiterjoin
\draw (0.0\du,15.0\du)--(0.0\du,16.5\du)--(6.0\du,16.5\du)--(6.0\du,15.0\du)--cycle;
\node at (3.0\du,15.8\du){linear};

\pgfsetmiterjoin
\draw (12.0\du,15.0\du)--(12.0\du,16.5\du)--(18.0\du,16.5\du)--(18.0\du,15.0\du)--cycle;

\node at (15.0\du,15.8\du){sigmoid};

\pgfsetbuttcap
{
\pgfsetarrowsend{latex}
\draw (5.0\du,19.0\du)--(1.0\du,16.5\du);
}

\pgfsetbuttcap
{
\pgfsetarrowsend{latex}
\draw (13.0\du,19.0\du)--(5.0\du,16.5\du);
}

\pgfsetbuttcap
{
\pgfsetarrowsend{latex}
\pgfsetstrokecolor{dialinecolor}
\draw (13.0\du,19.0\du)--(1.0\du,16.5\du);
}

\pgfsetbuttcap
{
\pgfsetarrowsend{latex}
\draw (5.0\du,19.0\du)--(5.0\du,16.5\du);
}

\pgfsetbuttcap
{
\pgfsetarrowsend{latex}
\draw (5.0\du,19.0\du)--(13.0\du,16.5\du);
}

\pgfsetbuttcap
{
\pgfsetarrowsend{latex}
\draw (13.0\du,19.0\du)--(17.0\du,16.5\du);
}

\pgfsetbuttcap
{
\pgfsetarrowsend{latex}
\draw (13.0\du,19.0\du)--(13.0\du,16.5\du);
}

\pgfsetbuttcap
{
\pgfsetarrowsend{latex}
\draw (5.0\du,19.0\du)--(17.0\du,16.5\du);
}

\pgfsetbuttcap
{
\pgfsetarrowsend{latex}
\draw (15.0\du,15.0\du)--(12.0\du,12.50\du);
}

\pgfsetbuttcap
{
\pgfsetarrowsend{latex}
\draw (15.0\du,15.0\du)--(15.0\du,12.50\du);
}

\pgfsetbuttcap
{
\pgfsetarrowsend{latex}
\draw (15.0\du,15.0\du)--(18.0\du,12.5\du);
}

\node at (12.0\du,11.6\du){$\alpha(0,0)$};

\node at (15.0\du,11.6\du){\dots};

\node at (18.5\du,11.6\du){$\alpha(T,F)$};

\end{tikzpicture}
	\endgroup
	\caption{Modified network architecture to better cope with background noise. It takes as input the log magnitude spectrogram of the mixture and has as output an embedding vector and a noise mask for each bin in the spectrogram.}
	\label{fig:deep_embedding_noise}
\end{figure}

Figure~\ref{fig:deep_embedding_noise} shows the modified network architecture. Besides an embedding vector, it now has a (scalar) mask output for each time-frequency bin. This scalar is an estimated ratio mask to suppress the noise in that bin. Because noise and speech signals have different roles and structures, there is no need for permutation invariance and the network can therefore directly output a noise filter mask.  
\subsubsection{Deep clustering}
\label{sec:deep_clustering_noise}
Loss function Eq.~(\ref{eq:deep_clustering_kost}) is modified to:
\begin{equation}
\sum_{n=1}^{N} \dfrac{1}{\tilde{K}_n^2} ||\tilde{\symbolmatrix{V}}_n\tilde{\symbolmatrix{V}}_n^T-\tilde{\symbolmatrix{Y}}_n\tilde{\symbolmatrix{Y}}_n^T||_F^2 + \gamma \frac{1}{K_n} ||\symbolvec{\alpha}_n -\symbolvec{\alpha}^{\text{ ideal}}_n ||_F^2
\end{equation}
with $K_n$ as defined previously, $\tilde{K}_n$ the number of bins not dominated by noise, $\tilde{\symbolmatrix{Y}}_n$ and $\tilde{\symbolmatrix{V}}_n$ as defined previously but the rows associated with bins dominated by noise are set to zero, $\symbolvec{\alpha}$ the ratio mask estimated by the network, and $\symbolvec{\alpha}^{ideal}$ the optimal ratio mask to filter the noise. The first term is similar to Eq.~(\ref{eq:deep_clustering_kost}). The second term trains the network to generate ratio masks to filter out the noise by penalizing the distance between the estimated and the optimal mask. The hyper-parameter $\gamma$ weighs the importance of separating the speakers and filtering out noise. In our experiments in Section~\ref{sec:noise_results} $\gamma$ is arbitrarily set to one. 

Also the procedure to separate unseen mixtures is modified. Firstly, before separating the speakers the estimated noise mask is used to suppress the noise. Secondly, only the embedding vectors for which are the associated $\alpha$ is greater than $0.75$ are used in the clustering algorithm. The remaining bins are assigned to one of the speakers based on the distance between their embedding vector and the cluster centres of the speakers. Based on these clusters, a binary mask to separate the speakers is generated.

\subsubsection{Deep attractor networks}
\label{sec:deep_attractor_noise}
For deep attractor networks the loss function Eq.~(\ref{eq:deep_attractor_networks_loss}) is modified to:
\begin{equation}
\sum_{n=1}^{N}\frac{1}{K_n*C_n}\sum_{c=1}^{C_n} ||\symbolmatrix{S}_{c,n}^{mag} - (\symbolmatrix{X}_n^{mag}\odot \symbolmatrix{M}_n^{noise}) \odot \symbolmatrix{M}_{n,c}||_{F}^2
\end{equation}
with $\symbolmatrix{M}^{noise}[t,f] = \alpha(t,f)$ the estimated noise mask. Also Eq.~(\ref{eq:dan_attractor_point}) is modified: 
\begin{equation}
\symbolvec{a}_c = \frac{\sum_{(t,f)}\symbolvec{v}(t,f) \tilde{y}_{(t,f),c}}{\sum_{(t,f)}\tilde{y}_{(t,f),c}}
\end{equation}
with $\tilde{y}_{(t,f),c}$ equal to one when speaker $c$ dominates the bin, the bin has enough energy and $\alpha(t,f)$ bigger than 0.75 and zero in all other cases. Although this hard cut-off introduces discontinuities and local optima in the cost function, an alternative (smoother) penalty for noisy bins did not lead to improved performance. 

To separate new mixtures, a similar strategy as in Section~\ref{sec:deep_attractor} is applied, but with two slight modifications. Firstly, prior to separating the speakers, the noise was filtered using the estimated noise mask. Secondly, to estimate the attraction points the K-means clustering is only applied to embedding vectors of time-frequency bins with enough energy and $\alpha$ above 0.75.
\subsection{Experiment set-up}
\label{sec:noise_design}
The utterances for the experiments in \ref{sec:noise_results} were sampled from the  `Wall Street Journal Database'\cite{wsj}. The noise signals were chosen from the `third CHiME speech separation and recognition challenge' data set \cite{chime3}, which contains recordings of realistic environment noise. As in the previous sections all signals were first downsampled to 8kHz. Six different two-speaker mixture sets were used:
\begin{itemize}
	\setlength{\itemsep}{1pt}
	\setlength{\parskip}{0pt}
	\setlength{\parsep}{0pt}
	\item A noise free training (20 000 mixtures) and development set (5 000 mixtures). The signals are normalised such that the individual speakers have the same power.
	\item A noisy training (100 000 mixtures) and development set (5 000 mixtures). The training set reuses each mixture of the noise free training set five times, each time with different noise. The new development set is similar to the noise free variant, only with noise added. The signals of the speakers and the noise are normalised such that they have the same power. 
	\item A noisy test set of 3 000 mixtures with different speakers and utterances than in the training and development sets. The noise comes from different parts of the same recordings as the training and development sets (for the training and development sets noise is sampled from the first 10 minutes of the recording, for the test set from the leftover part). The signals of the speakers and the noise are normalised such that they have equal power.
	\item A second noisy test set of 3 000 mixtures. Similar to the previous test set but now the signals are normalised such that both speakers have equal power and the noise is 3dB weaker than each speaker.
\end{itemize} 
\subsection{Results}
\label{sec:noise_results}
 Table \ref{tab:denoise_0db_3db} compares the performance of the following five methods for the two noisy test sets described in \ref{sec:noise_design}: 
\begin{itemize}
		\setlength{\itemsep}{1pt}
	\setlength{\parskip}{0pt}
	\setlength{\parsep}{0pt}
	\item deep clustering trained without noise (DC no noise);
	\item deep attractor networks trained without noise (DAN no noise);
	\item deep clustering with noise (DC with noise). During training the noise was considered as third speaker and the network was trained to form three clusters: two associated with speakers and one associated with the noise. During testing three reconstructions were created but only the two that most resembled a speaker were used for scoring; 
	\item modified deep clustering described in Section~\ref{sec:deep_clustering_noise} (modified DC);
	\item modified deep attractor networks described in Section~\ref{sec:deep_attractor_noise} (modified DAN).
\end{itemize}
For all methods a 20 dimensional embedding space was used. The recurrent part of the networks trained without noise consisted of two layers with 800 bidirectional LSTM cells each. For the networks trained with noise this consisted of four layers with each 800 bidirectional LSTM cells. 

The models trained without noise break down on noisy data. Including noise during training as a third speaker already leads to improved performance. The best SDRs are obtained with the modified methods of Section~\ref{sec:deep_embedding_noise}.
The SDR improvement w.r.t. ``DC with noise'' comes at a cost of a few dB in SNR, which seems less important since noise is not the main source of distortion. 

\begin{table}[th]
	\centering
	\caption{The average SDR and SNR in dB for the test sets with respectively the two speakers and the noise equally loud (0 dB) and the noise 3dB quieter than the speakers (3 dB)}
	\label{tab:denoise_0db_3db}
	\renewcommand*{\arraystretch}{1.2}
	\begin{tabular}{l cc cc}
		\toprule
		& \multicolumn{2}{c}{0 dB} & \multicolumn{2}{c}{3 dB} \\
		Method & SDR & SNR & SDR & SNR \\ \midrule
		DC no noise& -1.75 &  5.38 & 1.99  & 11.5\\
		DC with noise & 4.33 & 16.1 & 6.17 & 19.2 \\
		modified DC & 5.11 & 12.8  & 7.43 & 17.2 \\
		DAN no noise & -0.37 & 5.83 & 2.67 & 10.8 \\
		modified DAN & 5.27 & 13.5 & 7.33 & 17.4  \\ \bottomrule
	\end{tabular}
	
\end{table}
\section{Conclusion}
\label{sec:conclusion}
Deep clustering and deep attractor networks are applicable to source separation in a wide variety of languages, including tonal languages. Training models with (a combination of) related languages yields only minor performance degradation compared to training on the target language. 
This observation supports the results in \cite{zegers2018memory}, which showed that recurrent networks trained for speech separation mainly exploit information with the time span of a phone and long span information is limited to speaker identity while lexical or grammatical patterns are ignored.
Furthermore,we extended deep clustering and deep attractor networks with an estimated spectral mask to cope with noisy mixtures and showed significant improvement over the baselines. 
A limitation of the current experiments is that they only examine how well the methods perform for noise for which we have training data. Future work will consider ``untrained'' noise types. 

\bibliographystyle{IEEEtran}

\bibliography{references}


\end{document}